\documentclass{article}
\usepackage{spconf,amsmath,graphicx}

\title{
Identity-Sensitive Knowledge Propagation for Cloth-Changing Person Re-identification
}
\newcommand\authorsep{\quad}
\newcommand\lastauthorsep{\quad}
\name{Jianbing Wu$^\dagger$\authorsep Hong Liu$^{\dagger  *}$\authorsep Wei Shi$^\dagger$\authorsep Hao Tang$^\ddagger$
    \lastauthorsep Jingwen Guo$^\dagger$  
        \thanks{ $^*$Corresponding author. %
            This work is supported by National Natural Science Foundation of China (No.62073004), and Shenzhen Fundamental Research Program (No.JCYJ20200109140410340).
            }}
\address{
    $^\dagger$ Key Laboratory of Machine Perception, Shenzhen Graduate School, Peking University, China\\
    $^\ddagger$ Computer Vision Lab, ETH Zurich, Switzerland \\
    \fontsize{8.8}{10}\selectfont
    \texttt{\{kimbing.ng,jingwenguo\}@stu.pku.edu.cn, \{hongliu,pkusw\}@pku.edu.cn, hao.tang@vision.ee.ethz.ch}
}
\usepackage{enumitem}
\usepackage{lipsum}
\usepackage{textpos}

\usepackage{xspace}
\usepackage{multirow}
\usepackage{hhline}
\usepackage{diagbox} %
\usepackage{booktabs}
\usepackage{makecell}
\usepackage{mathtools}
\usepackage{blindtext}
\usepackage{nccmath} %
\usepackage[strict]{changepage} %
\usepackage{cite}
\usepackage{color}
\usepackage{mathrsfs}
\usepackage{amssymb}
\usepackage{algorithm}
\usepackage{algpseudocode}
\usepackage{balance}

\definecolor{dark}{RGB}{60, 65, 200}
\definecolor{pur}{RGB}{255, 60, 255}

\definecolor{dashblue}{RGB}{30,40,168}
\usepackage[colorlinks,
            linkcolor=dashblue,
            citecolor=green,
            ]{hyperref}
\usepackage[htt]{hyphenat} %
\usepackage[all]{hypcap}
\usepackage{cleveref}

\crefformat{figure}{Fig. #2#1#3}
\crefformat{table}{Table #2#1#3}
\crefformat{equation}{Formula #2(#1)#3}

\newcolumntype{L}[1]{>{\raggedright\arraybackslash}m{#1}}
\newcolumntype{C}[1]{>{\centering\arraybackslash}m{#1}}
\newcolumntype{R}[1]{>{\raggedleft\arraybackslash}m{#1}}

\newcommand{\vb}{\boldsymbol{b}}

\newcommand{\vf}{\boldsymbol{f}}

\newcommand{\vy}{\boldsymbol{y}}
\newcommand{\vz}{\boldsymbol{z}}

\newcommand{\MF}{\boldsymbol{F}}

\newcommand{\MW}{\boldsymbol{W}}
\newcommand{\MX}{\boldsymbol{X}}

\newcommand\ccreid{CC-ReID\xspace}
\newcommand\reid{Re-ID\xspace}
\newcommand\preid{Person re-identification\xspace}

\newcommand\fullmodelnamestrong{\textit{I\textbf{de}ntity-\textbf{S}ensitive \textbf{K}nowledge \textbf{Pro}pagation framework}\xspace}

\newcommand\modelnamenoxspace{DeSKPro}
\newcommand\modelname{\modelnamenoxspace\xspace}
\newcommand\modelnameplus{\modelnamenoxspace$^*$\space}

\newcommand\noclothstream{Mask-Guided Cloth-irrelevant Feature\xspace}
\newcommand\facestream{Facial Feature Enhancement\xspace}
\newcommand\globalstream{global stream\xspace}  %
\newcommand\localstream{face stream\xspace}  %

\newcommand\attnmodule{Cloth-irrelevant Spatial Attention\xspace}
\newcommand\attnmoduleshort{CSA\xspace}
\newcommand\celebreid{Celeb-reID\xspace}
\newcommand\celeblight{Celeb-reID-light\xspace}
\newcommand\prcc{PRCC\xspace}
\newcommand\fkploss{\loss_{fkp}}

\newcommand\metricsindicator{}

\usepackage{scalerel}

\newcommand\etal{\textit{et al.}\xspace}
\begin{document}
\maketitle
\begin{abstract}
\frenchspacing
Cloth-changing person re-identification (\ccreid), which aims to match person identities under clothing changes, is a new rising research topic in recent years.
However, typical biometrics-based \ccreid methods often require cumbersome pose or body part estimators to learn cloth-irrelevant 
features from human biometric traits, 
which comes with 
high computational costs.
Besides, the performance is significantly limited due to the resolution degradation of surveillance images. 
To address the above limitations,
we propose an effective \fullmodelnamestrong (\modelname) for CC-ReID.
Specifically,
a {\attnmodule} module %
is introduced
to eliminate the distraction of clothing appearance
by acquiring knowledge from the
human parsing module.
To mitigate the resolution degradation issue and 
mine identity-sensitive cues from human faces,
we propose to restore the missing facial details using prior facial knowledge,
which is then propagated
to a smaller network.
After training, %
the extra computations for human parsing or face restoration are no longer required.
Extensive experiments show that our framework outperforms state-of-the-art methods by a large margin.
Our code is available at \url{https://github.com/KimbingNg/DeskPro}.
 \end{abstract}
\begin{keywords}
\hspace{-0.59111pt}Cloth-Changing Person Re-Identification, Identity-Sensitive Features, Knowledge Propagation
\end{keywords}

\let\oldsection\section
\renewcommand\section[1]{
\vspace{-2.5pt}
\oldsection{#1}
\vspace{-2pt}
}
\let\oldsubsection\subsection
\renewcommand\subsection[1]{
\vspace{-2.5pt}
\oldsubsection{#1}
\vspace{-2pt}
}

\section{Introduction}

\noindent \preid (\reid),  which aims to 
match pedestrians 
across non-overlapping cameras, is a challenging task with significant research impact. %
Recent works \cite{liHarmoniousAttentionNetwork2018,sunPartModelsPerson2018,wangLearningDiscriminativeFeatures2018,herzogLightweightMultiBranchNetwork2021a} have achieved remarkable progress by learning powerful appearance representations 
based on the assumption that the images of the same identity in both the query set and the gallery set have the same clothing.
However, 
the clothing appearance tends to be 
unreliable in realistic scenes since people may wear different clothes at different times.

To address the above practical problem,
Cloth-Changing Person Re-identification (\ccreid) has drawn increasing attention in recent years
\cite{yuCOCASLargeScaleClothes2020,wanWhenPersonReidentification2020,huangScalarNeuronAdopting2020,yang2020clothing,Li_2021_WACV,xuAdversarialFeatureDisentanglement2021,Huang_2021_ICCV,Hong_2021_CVPR,shi2021iranet,shuSemanticGuidedPixelSampling2021}.  %
Existing 
\ccreid methods often focus on mining identity-relevant cues from pre-defined biometric traits.
\frenchspacing
For instance,
Yang \etal \cite{yang2020clothing} introduced a spatial polar transformation on contour sketch to learn shape representations.  %
Shi \etal \cite{shi2021iranet}
proposed to
learn head-guided features with the help of the pose estimator\cite{gulerDensePoseDenseHuman2018}.
Qian \etal \cite{qianLongTermClothChangingPerson2021} 
introduced a shape embedding module to extract biological features from body keypoints.
A common issue behind these methods is that  extra pose or contour estimation is required during inference, leading to higher computational costs. %
Some other researchers are dedicated to mining the identity-sensitive features from human faces\cite{yuCOCASLargeScaleClothes2020,wanWhenPersonReidentification2020} 
since facial cues are more robust 
under clothing changes.
However, due to the low resolution of  surveillance images, these methods also fail to extract representative facial features. %

\newcommand\smallbullet{\scaleto{\cdot}{6pt}}
\begin{figure}[tb]
  \centering
  \includegraphics[width=0.85\linewidth]{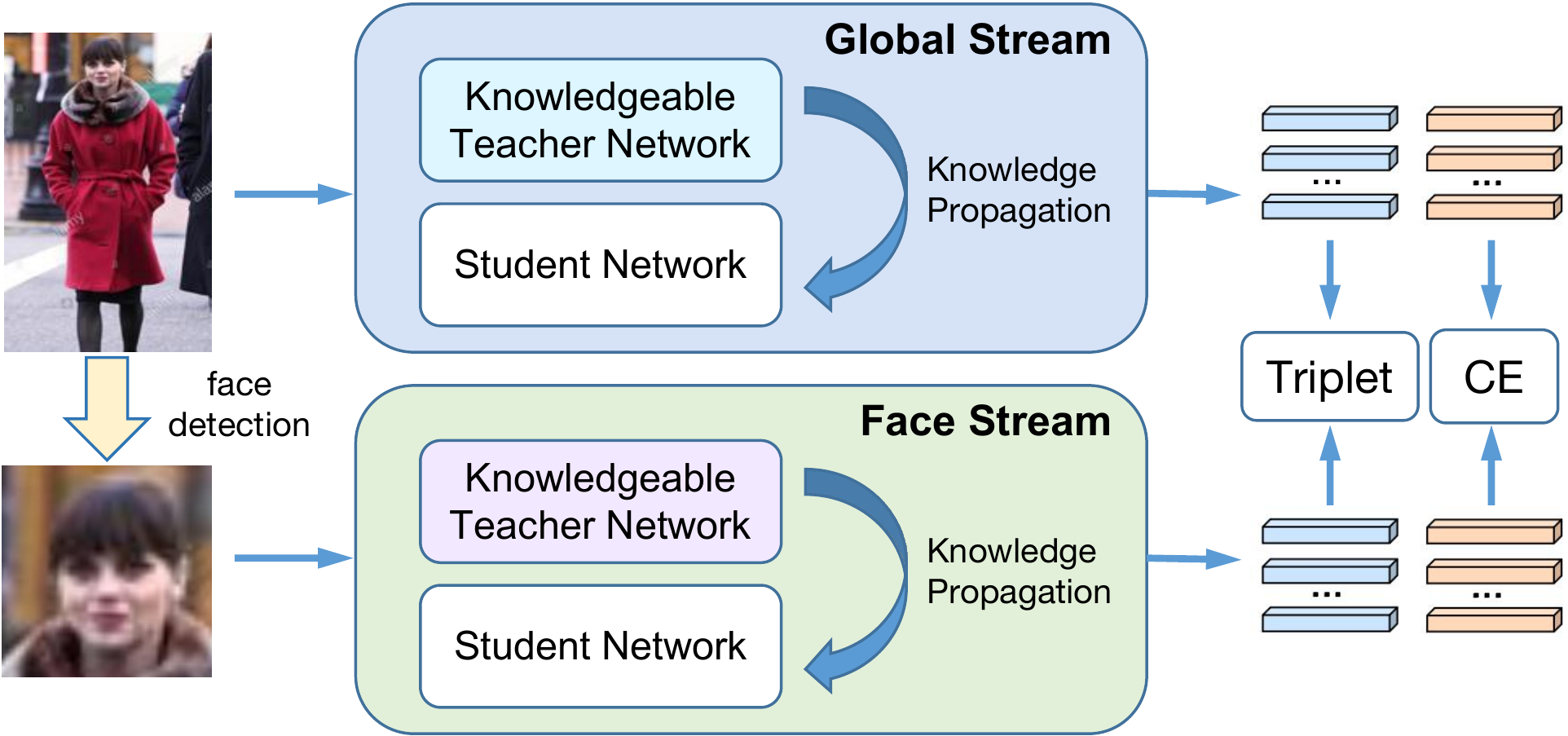}
  \vspace{-4pt}
  \caption{Overall architecture of the proposed framework. 
      ``CE" and ``Triplet" denote two widely used \reid losses, namely the cross-entropy loss and the triplet loss\cite{hermansDefenseTripletLoss2017}. %
  } 
  \vspace{-12pt}
  \label{fig:overall}
\end{figure}
\begin{figure*}[!htb]
  \centering
  \includegraphics[width=\linewidth]{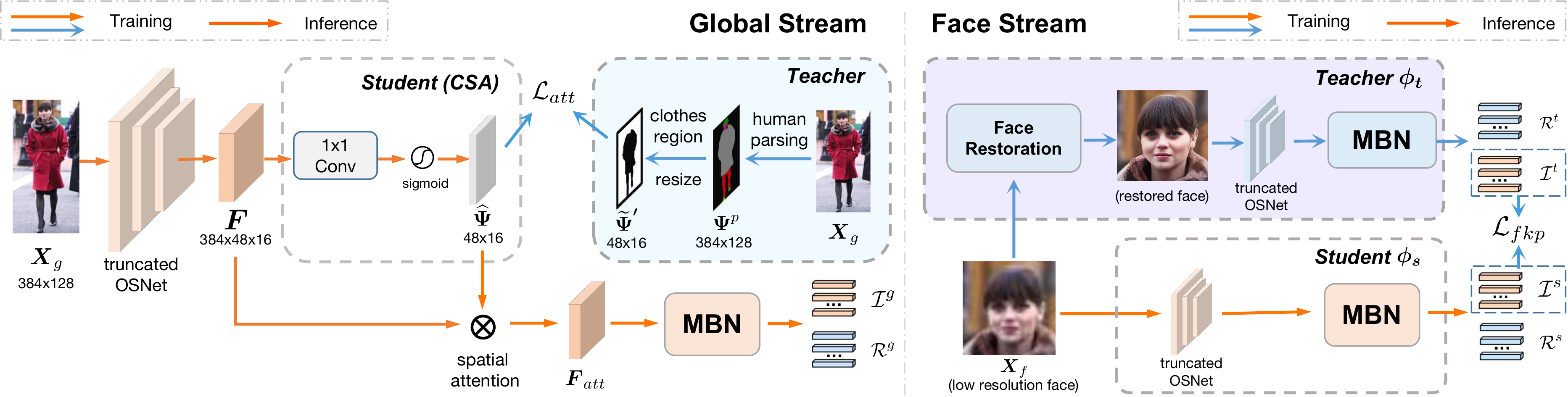}
  \caption{The architecture of the \globalstream and the \localstream.
The \globalstream aims to learn cloth-irrelevant knowledge under the guidance of the human parsing module,
and the \localstream is designed to extract representative facial features from low-resolution images by acquiring knowledge from the teacher network.
The details of notations can be referred to in \cref{sec:method}.
  } 
  \label{fig:architecture}
\end{figure*}

\frenchspacing
To address these challenges%
, we propose an %
\fullmodelnamestrong, termed \modelname.
As illustrated in \cref{fig:overall}, the proposed framework 
consists of a global \MakeLowercase{Cloth-irrelevant Feature} stream (\globalstream) and a \MakeLowercase{\facestream} stream (\localstream).
Both streams can acquire informative knowledge from their respective teacher networks during training.
In the \globalstream, 
a {\attnmodule} (\attnmoduleshort) module,  
which is trained under the guidance %
of the knowledgeable human parsing network,
is designed
to facilitate the learning of identity-sensitive features.
In the \localstream, %
we begin by training a teacher network, which can restore missing facial details from degraded images and learn representative features.
After that, a facial knowledge propagation loss is introduced to transfer the facial knowledge 
to a simpler student network. %
Note that with our well-designed knowledge propagation strategies, 
 the cumbersome  human parsing and the face restoration module can be removed during inference to 
reduce redundant computations.

Our contributions are summarized as follows:
\textbf{(1)} 
We introduce a \MakeLowercase{\attnmodule} module for the \ccreid task to effectively discourage the misleading features of clothes regions while preserving the identity-sensitive ones. %
\textbf{(2)} 
To mitigate the resolution degradation issue of surveillance images,
we further propose the idea of restoring the missing facial details and propagating the facial knowledge to a smaller network.
\textbf{(3)} 
Extensive experiments show that the proposed framework outperforms existing methods by a large margin without relying on the human parsing or face restoration module during inference.

\newcommand\globalinputimage{\MX_g}
\newcommand\faceimage{\MX_f}
\newcommand\feature{\MF}
\newcommand\loss{\mathcal{L}}
\newcommand\osnetout{\feature}
\newcommand\fattn{\feature_{att}}
\newcommand\mask{\boldsymbol{\Psi}}
\newcommand\attnmask{\widehat{\mask}}
\newcommand\convweight{\MW_{pw}^{}}
\newcommand\pointwiseconv[1]{\convweight * #1 + \vb}

\renewcommand{\paragraph}[1]{\vspace{2pt}\noindent\textbf{#1}.~}
\section{Methodology }\label{sec:method}

The key to addressing the \ccreid problem is to reduce the attention to the clothes regions and learn identity-sensitive features from biometric traits, such as human faces. 
We thus design a two-stream architecture, which consists of a global stream and a face stream, to learn both global cloth-irrelevant features and enhanced facial features. %
\cref{fig:architecture} illustrates the architecture of the two streams, and 
more details are discussed in the following subsections.

\subsection{\noclothstream Stream}\label{subsec:globalstream}
The clothing appearance is not reliable 
in cloth-changing settings.
A naive solution is to utilize spatial attention mechanisms
to explicitly discourage clothing features.
However, it is difficult to learn effective attention weights without auxiliary supervision.
Instead, we design a new {\attnmodule} (\attnmoduleshort) module to learn effective attention maps under the guidance of the human parsing network.

\paragraph{\attnmodule}
As illustrated in the \globalstream of \cref{fig:architecture},
given a person image $\globalinputimage$,
we first pass it through the truncated OSNet\cite{Zhou_2019_ICCV} backbone
{up until the first layer of its third convolutional block %
to derive the middle feature maps $\osnetout$ with size $h_f\times w_f\times c_f$.
Taking $\osnetout$ as the input, our \attnmoduleshort module then produces a cloth-irrelevant attention map, which can be formulated as:
\begin{equation}
\vspace{-2pt}
\attnmask = { \sigma \big(\pointwiseconv{\osnetout}\big)},
\vspace{-1pt}
\end{equation}
where $*$ denotes convolution operation, $\convweight$ denotes the point-wise convolution filters, and $\sigma(x) = 1/(1 + exp(-x))$ is the sigmoid function.
The attention map $\attnmask$ %
is then applied to the feature $\osnetout$ using
the following formula:
\begin{equation}
\fattn = \osnetout \otimes \attnmask,
\vspace{-1pt}
\end{equation}
where $\otimes$ denotes the Hadamard matrix product, and
$\fattn$
is the refined feature maps. %
We then feed $\MF_{att}$ into the Multi-Branch Network (MBN)
introduced in 
\cite{herzogLightweightMultiBranchNetwork2021a},
which combines global, part-based, and channel features in a multi-branch architecture. %
The resulting embeddings are grouped into two sets following \cite{herzogLightweightMultiBranchNetwork2021a}, denoted by $\mathcal{I}^g$ and $\mathcal{R}^g$, respectively.

\paragraph{Mask Guided Attention Loss} 
The attention map $\attnmask$ is expected to have lower scores in the clothes regions.
However, without extra auxiliary supervision, it is hard to guarantee that the \attnmoduleshort module produces cloth-irrelevant attention scores.
To this end, we introduce a mask-guided attention loss to enable the \attnmoduleshort module to learn effective attention maps.
As depicted in \cref{fig:architecture}, 
we first utilize the pre-trained human parsing module\cite{liSelfCorrectionHumanParsing2021} to estimate a fine-grained mask $\mask^{p}$ of human parts taking the image $\globalinputimage$ as the input.
Each grid in the mask $\mask^{p}$ represents the category of the corresponding pixel in $\globalinputimage$ (e.g., arm, leg, dress, skirt).
After that, 
a cloth-irrelevant mask can be obtained using the following formula:
\begin{equation}
\widetilde{\mask}^{}_{(i, j)} =
\begin{cases}
    \varepsilon, & ~ \text{if }~\mask^{p}_{(i, j)} \in \mathcal{C}, \\
    1,  & ~ \text{otherwise}.
\end{cases}
\end{equation}
Here, $\mathcal{C}$ denotes the set of cloth-related categories, and $\varepsilon$ is a hyper-parameter to avoid producing near-zero attention scores.
The mask guided attention loss is defined as:
\vspace*{-6pt}
\begin{equation}
\loss_{att}=\frac{1}{h_f\cdot w_f}\sum_{i=1}^{h_f} \sum_{j=1}^{w_f}\left(\attnmask_{(i, j)}-\widetilde{\mask}'_{(i, j)}\right)^{2},
\vspace{-6pt}
\end{equation}
where $\widetilde{\mask}'$ is obtained by resizing $\widetilde{\mask}$ to the same size as $\attnmask$. %

\vspace{-6pt}

\subsection{\facestream Stream}

The neural systems of human beings can perfectly identify persons by exploiting facial cues, even with resolution-degraded images.
This is thanks to their capability of recovering missing facial details.
Inspired by this, we design a teacher-student network in the \localstream to propagate facial knowledge from 
the knowledgeable teacher network $\phi_{t}$, which is trained beforehand, to the separate student network $\phi_{s}$.
Both the teacher and the student networks take the face image $\MX_f$ as input, which is detected from the image $\MX_g$ by the face detector\cite{dengRetinaFaceSingleShotMultiLevel2020}. %
As illustrated in \cref{fig:architecture},
the teacher first restores the facial details using the pre-trained face restoration module GPEN \cite{Yang2021GPEN}.
The enhanced face is then fed to the truncated OSNet and the MBN %
to derive 
the feature vector set $\mathcal{R}^t$ and the logit output vector set $\mathcal{I}^t$.
By optimizing the cross-entropy loss using $\mathcal{I}^t$ and the triplet loss using $\mathcal{R}^t$,  %
a knowledgeable teacher model can be obtained.

After training the teacher network, we fix its parameters %
and propagate the informative facial knowledge to the simpler student network $\phi_s$, where the cumbersome face restoration module is removed.
Similar to the teacher $\phi_{t}$, the outputs of the student $\phi_s$ are also grouped into two sets, denoted by $\mathcal{I}^s$ and $\mathcal{R}^s$, respectively.
\renewcommand\exp[1]{ e^{#1} }
Inspired by the idea of \cite{hinton2015distilling}, 
we design a facial knowledge propagation loss to transfer the knowledge to the student, which is defined as:
\newcommand\sm[1]{\mathcal{S}(#1)}
\newcommand\softmax[2]{
{\exp{\left(#1_{i} / #2\right)}}\hspace{-.51286pt}/{\sum_{j} \hspace{-1pt}\exp {\left(#1_{j} / #2\right)}}
}
\newcommand\kl[2]{
KL\left(\sm{#2, \tau}~\big\|~\sm{#1, \tau}\right)
}
\begin{gather}\label{eq:fkploss}
\fkploss = \tau^2\sum\nolimits_{i} \kl{\mathcal{I}^s_{(i)}}{\mathcal{I}^t_{(i)}},
\end{gather}
where 
$KL$ denotes Kullback–Leibler divergence
and $\tau$ is the temperature parameter of the softmax function $\mathcal{S}$.
Here, 
the softmax function $\mathcal{S}$, defined as
$
\sm{\vz, \tau}_{i}\!=\!\softmax{\vz}{\tau}
$,
is designed to produce knowledgeable soft labels for the student $\phi_t$.
The term $\tau^2$ in \cref{eq:fkploss} is adopted to guarantee that the magnitude of the computed loss remains unchanged for different 
values of $\tau$.
By optimizing $\fkploss$, the student network $\phi_s$ is endowed with the capability of extracting discriminative facial features from low-resolution images without relying on the face restoration module.

\vspace{-5pt}
\subsection{Training and Inference}
\vspace{-5pt}
\paragraph{Training}
Apart from the aforementioned losses,
we also optimize the {batch hard} triplet loss\cite{hermansDefenseTripletLoss2017} using all features in $\mathcal{R}^g \cup  \mathcal{R}^s$, and the cross-entropy loss  using all logit outputs in $\mathcal{I}^g \cup  \mathcal{I}^s$ for basic discrimination learning, formulated as:
\newcommand\faceidloss{\alpha \fkploss + (1 - \alpha) \loss_{ce}^s}
\begin{equation}
\begin{gathered}
\loss_{ce}^g = \sum\nolimits_{\hat{\vy} \in  \mathcal{I}^g } {CE}(\hat{\vy}, \vy) , \loss_{ce}^s = \sum\nolimits_{\hat{\vy} \in  \mathcal{I}^s } {CE}(\hat{\vy}, \vy),\\
\loss_{trip} = \sum\nolimits_{\vf       \in \mathcal{R}^g } {Trip} (\vf, \vy)+ \sum\nolimits_{\vf \in \mathcal{R}^s } {Trip} (\vf, \vy),
\end{gathered}
\end{equation}
where $CE$ and $Trip$ denote the cross-entropy loss and the {batch hard} triplet loss, respectively.
The term $\vy$ denotes the identity labels of person images.
The overall framework is trained
by optimizing the total loss $\loss$, which is defined as:
\begin{equation}
\begin{gathered}
\loss = \lambda\loss_{att} + \loss_{trip} + \left(\faceidloss\right)  + \loss_{ce}^g,
\end{gathered}
\end{equation}
where the $\lambda$ and  $\alpha$ are adopted to balance different losses.

\paragraph{Inference} %
The inference stage,
during which the human parsing module in the \globalstream and the teacher network $\phi_t$ in the \localstream are removed,
can be seen as a cosine-similarity-based retrieval process.
Embeddings obtained before the last fully connected layer in the MBNs are concatenated for similarity computation as in \cite{herzogLightweightMultiBranchNetwork2021a}.
As for those query images with no face detected, we only 
enable the \globalstream to compute feature vectors.

\section{Experiments and Discussions}

\subsection{Datasets and Experimental Setups}

Our experiments are conducted on
three widely used 
\ccreid datasets, including \celebreid\cite{huangScalarNeuronAdopting2020}, \celeblight\cite{huangScalarNeuronAdopting2020}, and \prcc\cite{yang2020clothing}.
The \celebreid dataset is made up of person images where over 70\% of samples show different clothes.
The \celeblight dataset is the light but challenging version of \celebreid since all images of each person are in different clothes.
The \prcc dataset provides both cross-clothes and same-clothes settings to support in-depth evaluations.
The values of $\varepsilon$, $\lambda$, $\tau$, and $\alpha$ are determined by cross validation.
Specifically, 
for \celebreid and \celeblight, 
    we set $\varepsilon=0.1$, $\lambda=7$, $\tau=5$, and $\alpha=0.7$.
For \prcc dataset, 
    we set $\varepsilon=0.1$, $\lambda=7$, $\tau=1$, and $\alpha=0.8$.
Other 
experimental settings follows 
our previous work\cite{shi2021iranet}. %
The cumulative matching characteristics (CMC) and mean Average Precision (mAP) 
are reported in the following subsections.

\subsection{Comparison with State-of-the-Art Methods}
We compare our method with several common \reid models\cite{liHarmoniousAttentionNetwork2018,sunPartModelsPerson2018,wangLearningDiscriminativeFeatures2018,herzogLightweightMultiBranchNetwork2021a} 
and state-of-the-art \ccreid approaches\cite{huangScalarNeuronAdopting2020,yang2020clothing,Li_2021_WACV,xuAdversarialFeatureDisentanglement2021,Huang_2021_ICCV,Hong_2021_CVPR,shi2021iranet,shuSemanticGuidedPixelSampling2021}.
In the tables hereafter,
``R-$k$" denotes rank-$k$ accuracy, %
``-" denotes not reported, and the best and second-best results are in bold and underline styles, respectively.

\cref{tb:perfceleb} shows the experimental results on \celeblight and \celebreid datasets.
Compared with the state-of-the-art method IRANet\cite{shi2021iranet},
which relies on the off-the-shelf pose estimator\cite{gulerDensePoseDenseHuman2018},
our method achieves higher performance in both datasets.
For \prcc dataset, %
we conduct experiments
in both cross-clothes and same-clothes settings.
As shown in \cref{tb:perfprcc}, 
the proposed \modelname
outperforms
the best existing method
by a large margin in the cross-clothes setting.
Notice that our method surpasses FSAM \cite{Hong_2021_CVPR} as well, which also attempts to 
transfer knowledge from pre-trained networks to save extra computations.
This is because
they only complement shape knowledge in the appearance stream while omitting the identity-sensitive cues from human faces.
Moreover, 
the clothing appearance 
is not explicitly suppressed in their method, which damages the robustness of features.
Although \modelname does not outperform all existing methods in the same-clothes setting, it still attains competitive results.

\vspace{-3pt}
\subsection{Effectiveness of Proposed Components}
\vspace{3pt}
To verify the effectiveness of each component in \modelname, 
we conduct ablation experiments with different component settings on the more challenging \celeblight dataset.
\newcommand\bld[1]{\textbf{#1}}
\newcommand\udl[1]{\underline{#1}}

\begin{table}[tb]
\centering
\caption{Comparison on \celeblight and \celebreid datasets (\%).  }
\label{tb:perfceleb}
\vspace{3pt}
\resizebox{\linewidth}{!}{
\begin{tabular}{c|ccc||ccc}
        \toprule
        \multirow{ 2}{*}{~~~Method~~~} & \multicolumn{3}{c||}{\multirow{1}{*}{\text { \celeblight }}} & \multicolumn{3}{c}{\multirow{1}{*}{ \celebreid }} \\
\hhline{~------}
\hhline{~------}
& \text { R-1\metricsindicator } & \text { R-5\metricsindicator } & \text { mAP\metricsindicator } & \text { R-1\metricsindicator } & \text { R-5\metricsindicator } &  mAP\metricsindicator \\
\hhline{-------}
HACNN  \cite{liHarmoniousAttentionNetwork2018}            &    16.2    &     -      &    11.5    &    47.6    &     -      &    9.5    \\ 
MGN \cite{wangLearningDiscriminativeFeatures2018}         &    21.5    &     -      &    13.9    &    49.0    &     -      &    10.8   \\
AFD-Net \cite{xuAdversarialFeatureDisentanglement2021}    &    22.2    &    51.0    &    11.3    &    52.1    &    66.1    &    10.6   \\
LightMBN \cite{herzogLightweightMultiBranchNetwork2021a} &    32.9    &    67.0    &    18.8    &    57.3    &    71.6    &    14.3   \\
ReIDCaps+ \cite{huangScalarNeuronAdopting2020}            &    33.5    &    63.3    &    19.0    &    63.0    &    76.3    &    15.8   \\
RCSANet \cite{Huang_2021_ICCV}                            &    \udl{46.6}    &     -      &    24.4    &    65.3    &     -      &    17.5   \\
CASE-Net \cite{Li_2021_WACV}                              &    35.1    &    66.7    &    20.4    &    \udl{66.4}    &    78.1    &    18.2   \\
IRANet \cite{shi2021iranet}                               &    46.2    &    \udl{72.7}    &    \udl{25.4}    &    64.1    &    \udl{78.7}    &    \udl{19.0}   \\
\hline
\textbf{\modelname (Ours)}                                &    \textbf{52.0}    &    \textbf{81.6}    &    \textbf{29.8}    &    \textbf{68.6}    &    \textbf{82.3}    &    \textbf{22.7}   \\
\bottomrule
\end{tabular}
}
\end{table}

\begin{table}[tb]
\centering
\vspace{-10pt}
\caption{Comparison on PRCC dataset (\%).}
\label{tb:perfprcc}
\vspace{3pt}
\resizebox{\linewidth}{!}{
\newcommand\customwidth{1.2cm}
\begin{tabular}{C{2.8cm}|C{\customwidth}C{\customwidth}|C{\customwidth}C{\customwidth}}
\toprule
\multirow{ 3}{*}{~~~Method~~~} & \multicolumn{4}{c}{\text { PRCC }} \\
\cline { 2-5 } & \multicolumn{2}{c|}{Cross clothes} & \multicolumn{2}{c}{ Same clothes} \\
\cline { 2-5 }  &  \text { R-1\metricsindicator }&  \text { mAP\metricsindicator } & \text { R-1\metricsindicator }& \text { mAP\metricsindicator } \\
\hhline{-----}
HACNN  \cite{liHarmoniousAttentionNetwork2018}            &    21.8    &     -      &    82.5    &     -     \\
PCB \cite{sunPartModelsPerson2018}                        &    22.9    &     -      &    86.9    &     -     \\
Yang \etal \cite{yang2020clothing}                       &    34.4    &     -      &    64.2    &     -     \\
CASE-Net \cite{Li_2021_WACV}                              &    39.5    &     -      &    71.2    &     -     \\
AFD-Net \cite{xuAdversarialFeatureDisentanglement2021}    &    42.8    &     -      &    95.7    &     -     \\
RCSANet \cite{Huang_2021_ICCV}                            &    50.2    &    48.6    &   \textbf{100.0}    &    97.2   \\
LightMBN \cite{herzogLightweightMultiBranchNetwork2021a} &    50.5    &    51.2    &   \textbf{100.0}    &    \bld{98.9}   \\
FSAM \cite{Hong_2021_CVPR}                                &    54.5    &     -      &    98.8    &     -     \\
IRANet \cite{shi2021iranet}                               &    54.9    &    53.0    &    \udl{99.7}    &    \udl{97.8}   \\
Shu \etal \cite{shuSemanticGuidedPixelSampling2021}      &    \udl{65.8}    &    \udl{61.2}    &    99.5    &    96.7   \\
\hline
\textbf{\modelname (Ours)}                                &    \textbf{74.0}    &    \textbf{66.3}    &    99.6    &    96.6   \\
\bottomrule
\end{tabular}
}
\end{table}

\paragraph{\attnmodule} 
We first evaluate the importance of the attention module \attnmoduleshort and the mask-guided attention loss $\loss_{att}$.
As shown in \cref{tb:ablation}, the rank-1 accuracy and mAP can be improved 
with the \attnmoduleshort module.
After adding $\loss_{att}$ to the total loss function, the 
performance
can be further boosted.
The feature visualization is presented in \cref{fig:visualization}, which also shows that the \attnmoduleshort model can effectively attend to the cloth-irrelevant regions after training with $\loss_{att}$.

\paragraph{Face Enhancement and Knowledge Propagation} 
    As shown in \cref{tb:ablation},
the teacher network (model 6) significantly outperforms the student network (model 4),
which verifies that restoring face details can help improve performance.
Comparing model 5 with model 4, it can be seen that the performance is boosted significantly after the knowledge propagation.
Even though model 5 
is still slightly inferior to
model 6,
it has lower computational costs since the face restoration module is removed.
This also indicates that the knowledge propagation strategy in the \localstream is the trade-off between performance and computational efficiency.

\begin{table}[tb]
\newcommand\globalbaseline{$\mathcal{G}$}
\caption{Ablation study on \celeblight dataset. 
``\globalbaseline" denotes the \globalstream without the \attnmoduleshort module,
    and $\phi^{+}$/$\phi^{-}$ denotes the student network trained with/without $\loss_{fkp}$.
}\label{tb:ablation}
\vspace{3pt}
\resizebox{\linewidth}{!}{
\begin{tabular}{c|ccc|ccc||C{1cm}C{1cm}}
\toprule
\multirow{2}{*}{Model} & \multicolumn{3}{c|}{Body Stream} & \multicolumn{3}{c||}{Face Stream}   & \multicolumn{2}{c}{\celeblight}                   \\
 \cline{2-9} 
 & \globalbaseline & \attnmoduleshort & \multicolumn{1}{c|}{$\loss_{att}$} & $\phi_s^-$   & $\phi_s^+$ & \multicolumn{1}{c||}{$\phi_t$} & \multicolumn{1}{c}{R-1\metricsindicator} & \multicolumn{1}{c}{mAP\metricsindicator} \\ 
 \hline
1 (baseline)            &  \checkmark         &            &            &            &            &            &    32.9    &    18.8   \\
2                       &  \checkmark     & \checkmark &            &            &            &            &    33.5    &    20.1   \\
3                       &  \checkmark     & \checkmark & \checkmark &            &            &            &    37.2    &    20.3   \\
\hline
4                       &       &            &            & \checkmark &            &            &    41.6    &    20.7   \\
5                       &       &            &            &            & \checkmark &            &    47.0    &    25.6   \\
6                       &       &            &            &            &            & \checkmark &    47.4    &    25.8   \\
\hline
7                       &  \checkmark     & \checkmark & \checkmark & \checkmark &            &            &    50.1    &    27.4   \\
\textbf{\modelname$~$}  &  \checkmark       & \checkmark & \checkmark &            & \checkmark &            & \udl{52.0} & \udl{29.8}\\
\textbf{\modelnameplus} &  \checkmark       & \checkmark & \checkmark &            &            & \checkmark & \bld{53.7} & \bld{30.9}\\
\bottomrule
\end{tabular}
}
\end{table}
 \begin{figure}[tb]
  \centering
  \includegraphics[width=.95\linewidth]{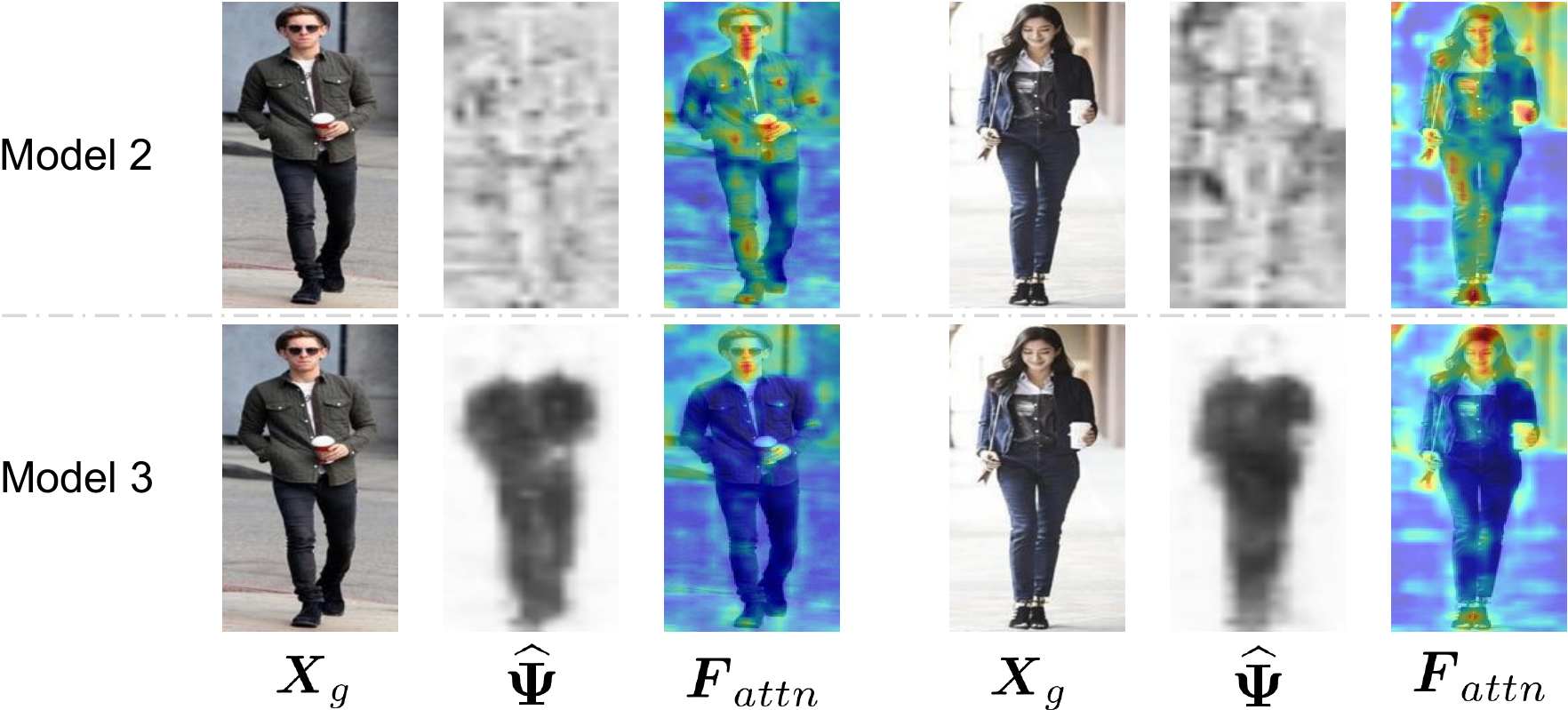}
  \vspace{-8pt}
  \caption{Visualization of the attention maps and the refined feature maps for model 2 and model 3.} 
  \label{fig:visualization}
\end{figure}

\paragraph{Two Stream Architecture of \modelname}
As shown in \cref{tb:ablation},
the performance can be further improved
by assembling both global and face streams.
This indicates that 
the global and the facial features complement each other.
We also provide a variant of \modelname, which is termed \modelnameplus in \cref{tb:ablation}.
It is obtained by replacing the student network $\phi_s$ with the more complex teacher network $\phi_t$ in the \localstream.
Even though it achieves the best performance in our experiments, 
     it brings extra computational costs for face restoration.
\section{Conclusion}
In this paper, we present an \fullmodelnamestrong (\modelname),
which 
achieves state-of-the-art results on three challenging \ccreid datasets, 
including \celeblight, \celebreid, and \prcc.
We argue that 
suppressing the clothes features explicitly 
and 
recovering facial details from resolution-degraded images 
can help boost the performance of the \ccreid task.
The experiments have also shown that propagating body and facial knowledge can help avoid the extra computation costs for mask estimation or face restoration while preserving performance.
We hope this work can spark further research on the \ccreid problem.

\clearpage
\nonfrenchspacing
\balance
\bibliographystyle{IEEEbib}

\end{document}